\documentclass[10pt,letterpaper]{article}

\usepackage{ccn}
\usepackage{pslatex}
\usepackage{apacite}
\usepackage{hyperref}
\usepackage{natbib}
\usepackage{lineno}

\usepackage{graphicx}
\usepackage{amsmath}
\usepackage{amsfonts} 
\usepackage{todonotes}
\usepackage{placeins}
\usepackage{pifont}

\title{Learning to Chain Operations by Routing Information \\Through a Global Workspace}
 
\author{
{\large \bf Hugo Chateau-Laurent (hugo.chateaulaurent@gmail.com)} \\
  CerCo, CNRS UMR 5549, Université de Toulouse\\
  \AND {\large \bf Rufin VanRullen (rufin.vanrullen@cnrs.fr)} \\
  CerCo, CNRS UMR 5549, Université de Toulouse and ANITI, Artificial and Natural Intelligence Toulouse Institute}

\begin{document}
\nolinenumbers

\maketitle
\section{Abstract}
{
\bf
We present a model inspired by the Global Workspace Theory that integrates specialized modules to perform a sequential reasoning task. A controller selectively routes information between modules through the workspace using a gating mechanism. This approach allows the model to chain operations by iteratively broadcasting information between specialized domains, mimicking System-2 reasoning. We evaluate the model’s performance on a simple addition task, where two addends must be summed. The task can be solved by routing information sequentially through an Input module, an Increment module (multiple times), and finally an Output module. We consider two implementations of this system with increasing complexity. First, using hand-designed modules operating on one-hot digit representations, the controller (a LSTM recurrent network) learns to select the appropriate modules (input, increment, output) in the appropriate sequence. Second, we replace the hand-designed modules with learned representation modules for MNIST images and an increment module trained on the task objectives; here again, the controller learns the appropriate sequential module selection to solve the task. Finally, we show that the Global Workspace model, while having fewer parameters, outperforms LSTMs and Transformers when tested on unseen addition operations (both interpolations and extrapolations of addition operations seen during training). Our results highlight the potential of architectures inspired by the Global Workspace Theory to enhance deep learning's reasoning capabilities.
}
\begin{quote}
\small
\textbf{Keywords:} 
Global Workspace; sequential reasoning; System 2; Transformer; LSTM, length generalization
\end{quote}

\section{Introduction}

In problem solving, individuals often rely on multiple strategies, beginning with slower deliberate processes and transitioning to faster automated responses as they gain experience. In the early stages of learning addition, children frequently use multi-step strategies, incrementing by one multiple times until reaching the final sum. This reflects System-2 reasoning \citep{kahneman2011thinking}, where each operation is performed sequentially. Over time, this method evolves to a System-1-like recall, where answers are retrieved more automatically from memory \citep{siegler1987perils}. While memorization yields faster responses, it is also more error prone and less flexible.

This distinction in cognitive science between sequential and automatic processing mirrors a current challenge in artificial intelligence: how to imbue models with System-2 reasoning capabilities. Most deep learning models, such as Long Short-Term Memory networks (LSTM) \citep{hochreiter1997long} and Transformers \citep{vaswani2017attention}, excel at System-1-like tasks where memorization and statistical correlations dominate \citep{bengio2019system}. They are also able to solve some arithmetic problems \citep{zaremba2014learning}. However, these architectures tend to falter in tasks that require stepwise reasoning, where multiple operations must be chained sequentially, such as incrementing numbers until reaching a final sum. An insightful demonstration has been provided that Transformers, LSTMs and convolution-based neural networks all struggle in length generalization, that is, to generalize to sequences that are longer than those seen during training \citep{hupkes2020compositionality}.

Previous attempts at endowing neural networks with System-2 abilities include granting them with external memory \citep{graves2014neural} or world models \citep{ha2018world}. A compelling alternative is offered by Global Workspace Theory (GWT) \citep{baars1993cognitive}, which describes how human cognition integrates information across specialized domains through conscious processing. In this theory, different cognitive modules compete for access to a global workspace, where the most relevant information is broadcasted and integrated for higher-level reasoning. \cite{sackur2009cognitive} studied the human execution of arithmetic tasks and highlighted that, while unconscious processes can handle isolated operations, chaining multiple steps in a sequence requires conscious control to ensure proper information flow. This alternative to standard machine learning architectures is largely uncharted (but see \cite{devillers2024semi, juliani2022perceiver, goyal2021coordination}). 

In this work, we follow the roadmap proposed by \cite{vanrullen2021deep} to grant deep learning architectures higher-level cognitive functions by chaining operations from specialized domains, using principles inspired by GWT. The model we present here captures some basic aspects of GWT via a simplified attentional mechanism. It contrasts sharply with the implicit black-box processing of LSTMs and Transformers; yet we demonstrate significant advantages in the acquisition of an addition task, and in its generalization to novel test conditions (interpolation and extrapolation).  

\section{Task}

\begin{figure*}[!tb]
\begin{center}
\includegraphics[clip, trim=0cm 12.7cm 9cm 0cm, width=\textwidth]{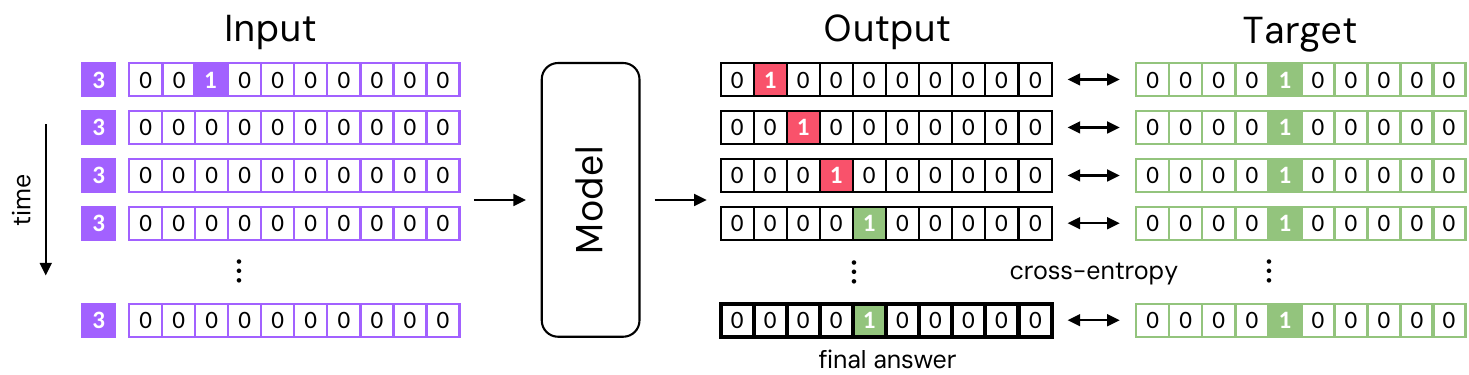}
\end{center}
\caption{Example input, output, and target. The input is the left addend (three in the example) and the one-hot encoding of the right addend (two in the example; in our second, MNIST model, this one-hot input is replaced by an image input from the MNIST dataset). The right addend is only shown in the first time step. The model outputs predicted sums (again, as a one-hot vector), which are compared to the target.} 
\label{fig:task}
\end{figure*}

The task we consider in this work is addition, where the goal is to compute the sum of two numbers, known as addends (Figure \ref{fig:task}). In our experiments, the left addend is \( A_L \in \mathbb{N}^+ \) and the right addend is \( A_R \in \mathbb{N} \cap [0,9]\). The right addend is only shown in the first time step. Although the left addend is unbounded, the models are only exposed to values of \( A_L \) between 1 and 9 during training (or subsets of these values for out-of-distribution generalization tests, see below). The answer to the problem is the modulo 10 sum of both addends: \( S = (A_L + A_R) \text{ mod } 10 \). This allows us to use a one-hot representation for the right addend (or an image from the MNIST dataset in our second model) as well as for the task output (the left addend being represented as a scalar).

A cross-entropy loss, averaged over multiple time steps (we use \( T=20 \)), is used to compare the model’s predicted probability distribution with the true one-hot encoded distribution. The model is thus encouraged to predict the target sum in as few time steps as possible. We evaluate accuracy on the model's output averaged over time steps.
 



\section{Global Workspace Model}

\begin{figure*}[!tb]
\begin{center}
  \includegraphics[clip, trim=1.7cm 12.2cm 10.54cm 0cm, width=\textwidth]{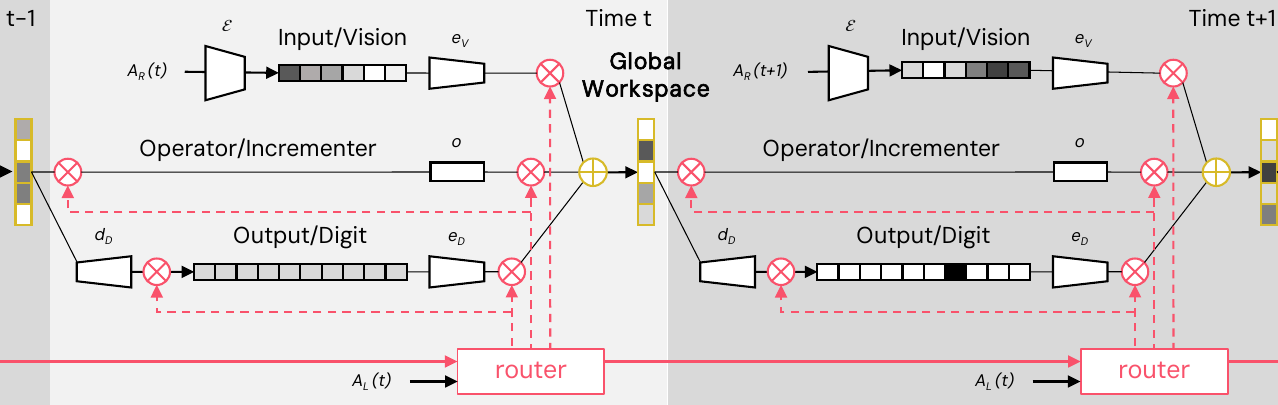}
  \end{center}\caption{Architecture of the Global Workspace model. In the one-hot model, encoders and decoders are set to the Identity, while the Operator/Incrementer module's weight matrix is set manually. In the MNIST model, input images $A_R(t)$ are encoded via a pretrained VAE $\mathcal{E}$; Global Workspace encoders and decoders are trained for multimodal representation objectives, and the Operator/Incrementer is trained from scratch to update the current GW representation given a task instruction $A_L=1..9$ and a ground-truth (``oracle'') routing sequence. For both models, the router (in red) is finally trained to produce a sequence of ``gates'' for the three modules. Given a task instruction $A_L=n$, the expected gating sequence (after learning) is to first select the input module to read out a digit in the GW; then call the Operator module $n$ times in a row to perform the addition and update the GW, before selecting the output module to return the GW representation as a one-hot digit answer.}
\label{fig:architecture}
\end{figure*}

The architecture we propose (Figure \ref{fig:architecture}) is inspired by GWT, which suggests that specialized modules compete for access to a shared workspace, the content of which can be broadcasted to all modules. Our model leverages this idea to integrate such modules for solving tasks that require sequential reasoning, like arithmetic operations. Each module of the model is responsible for a specific task, such as perceiving the right addend, performing increments, or outputting the final sum. These three modules do not operate independently. Instead, information flows between them through the central global workspace, which acts as a communication hub. An attentional mechanism is used to selectively route information between the workspace and the three modules, ensuring that the correct operations are performed in sequence. This router allows the model to chain multiple operations together, mimicking the step-by-step processing characteristic of System-2 reasoning. The router is modeled as a LSTM, which takes as input the left addend \( A_L \) and outputs a 3-dimensional vector $g(t)$ representing the interaction gates for each specialized module, which is then softmaxed. These gates control how information flows between the modules and the global workspace.

The modules can be divided into two domain modules (vision and digit) and an operation module (incrementer). Let \( V(t) \) and \( D(t) \) respectively represent the state of the vision and digit domains. The vision domain plays the role of the input module, as it encodes \( A_R \) in the first time step:
\begin{align}
    V(t) &= 
    \begin{cases}
        \mathcal{E}(A_R) & \text{if } t=0 \\
        0 & \text{otherwise}
    \end{cases}
\end{align}
where \( \mathcal{E} \) is the identity function when \( A_R \) is provided as a one-hot vector ($\text{Label}(A_R)$), and the encoder of a pretrained and frozen variational autoencoder when \( A_R \) is an MNIST image ($\text{Image}(A_R)$), as described later. The state of the digit domain, which is the output module from which the answer is read, is computed as follows:
\begin{align}
    D(t) &= g_D(t) d_D (gw(t-1))
\end{align}
where \( gw(t-1) \) is the previous global workspace state, \( g_D(t) \) is the gate for the digit module and $d_D$ is a feedforward network used by the digit domain to decode the workspace content.

The operation module (incrementer) does not interact with the external environment directly, but instead performs an operation \( o \) (also using a feedforward network) on the state of the global workspace. Therefore, it does not need an encoder nor a decoder:
\begin{align}
    O(t) &= o(g_O(t) gw(t-1))
\end{align}
The state of the global workspace is the aggregation of the three potential modules, i.e. the encoded domain states and the output of the increment operator, each of them weighted by the gate values given by the router:
\begin{align}
    gw(t) &= g_V(t) e_V(V(t)) + g_O(t) O(t) + g_D(t) e_D(D(t))
\end{align}

\section{Results}
\subsection{One-hot Model}
\begin{figure}[!tb]
\begin{center}
  \includegraphics[clip, trim=0cm 1.3cm .82cm .4cm, width=0.37\textwidth]{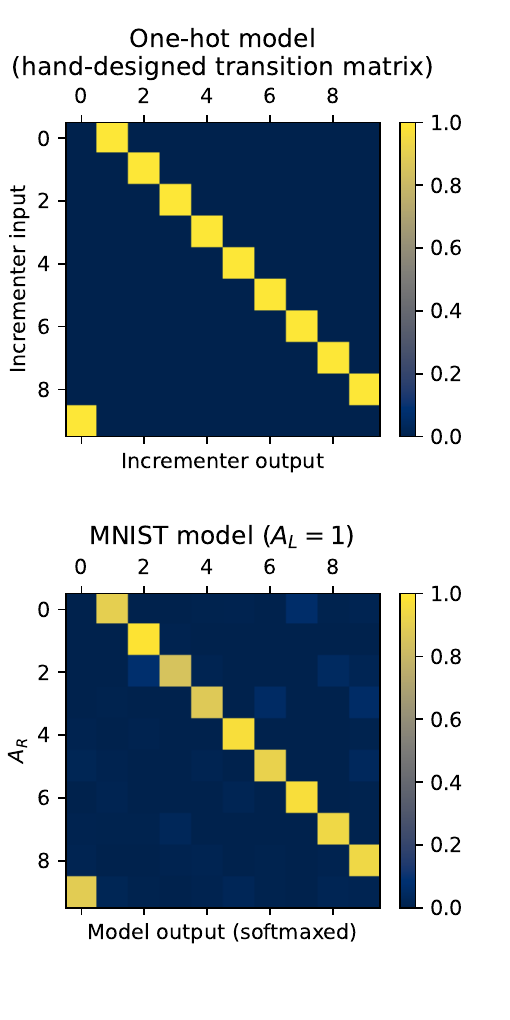}
  \end{center}
    \caption{Comparison between the hand-designed incrementer of the one-hot model (top) and the behavior of the MNIST model with $A_L=1$ (bottom). The bottom panel represents the softmaxed output (activity of the digit domain) at $t=2$ (i.e. after three routing steps). The operators of both models shift the digit by one in their respective representation.}
\label{fig:test_router}
\end{figure}

We start with a very simple version of the task, in which $A_R$ is represented as a 10-dimensional one-hot vector, and all modules (except the router) are hand-designed to process these vectors. The dimension of the modules is thus set to 10. $\mathcal{E}$, $d_D$, $e_V$ and $e_D$ are set to the identity  function, so they simply pass on the one-hot vector to the next stage. For simplicity, we also define $o$ as a circular shift function, such that each element of the one-hot vector is shifted right by one, with the last element wrapping around to the first position (see Figure~\ref{fig:test_router}, top).

We train the router LSTM while keeping other parameters of the Global Workspace model frozen. We optimize the test accuracy (after 10,000 epochs) with respect to the hyperparameters of the model, using bayesian search, while it is trained with a random 80\% split of the tasks.
The resulting hyperparameter values are reported in Supplementary Table \ref{table:hyperparameters} and used in the experiments.

\begin{figure}[!tb]
\begin{center}
  \includegraphics[clip, trim=0.5cm 2.1cm 0cm 2.6cm, width=0.5\textwidth]{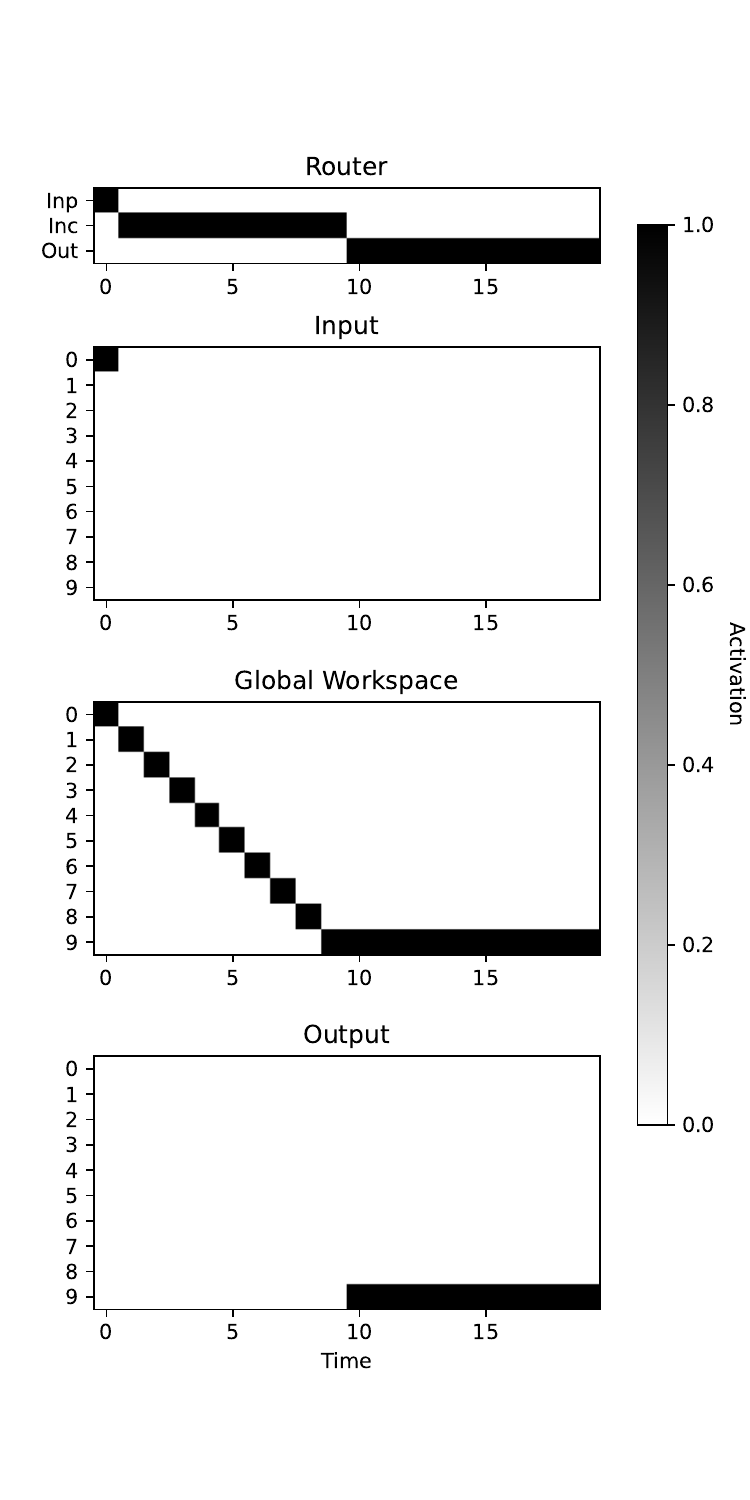}
  \end{center}
    \caption{Example behavior of the Global Workspace model adding \( A_L=9 \) to \( A_R=0 \). The top panel displays the three output gates of the Router over the simulation time steps. The next panel shows the one-hot encoded input (only available at time step 0). The third panel is the state of the global workspace (one-hot encoded). The last panel reflects the Output module (one-hot encoding). The Router appears to sequentially select the appropriate modules: first the Input, then the Increment module for 9 successive steps (allowing the global workspace representation to increase), and finally the Output producing the correct answer.}
\label{fig:states_9}
\end{figure}

Inspection of the model's behavior reveals that it has learned to correctly execute operations in the expected sequence (Figure \ref{fig:states_9}). This is true for all combinations of left and right addends in the test set ($(A_L,A_R) \in [1..9]_{\mathbb{N}}\times[0..9]_{\mathbb{N}}$): the test accuracy is $100\%$.

\subsection{MNIST Model}

We now turn to describing a more advanced version of the model that can accept MNIST images and one-hot encoded digits as input and output, respectively. We first pretrain a variational autoencoder on MNIST images, providing us with an image encoder $\mathcal{E}$ and decoder $\mathcal{D}$ that are kept frozen in the subsequent steps. The encoder comprises 3 ReLU convolutional layers with batch normalization followed by a feedforward layer providing us with a $12$-dimensional hidden state. Similarly, the decoder is composed of a feedforward layer followed by 3 deconvolutional layers with batch normalization and ReLU activation. The autoencoder was trained during $55$ epochs with a learning rate of $3\text{e}^{-4}$ and a batch size of $32$.

\begin{figure}[!tb]
\begin{center}
  \includegraphics[clip, trim=0cm 13.48cm 30.2cm 0cm, width=0.32\textwidth]{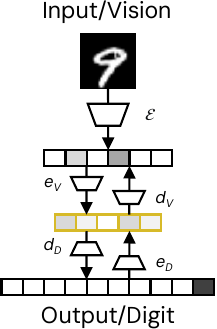}
  \end{center}
    \caption{Architecture of the multimodal global workspace. The MNIST images are encoded using $\mathcal{E}$. Furthermore, both the vision and digit domains can send information to the workspace through their encoder $e$, and read from it using their decoder $d$.}
\label{fig:architecture_shimmer_2}
\end{figure}

We then build upon the work of \cite{devillers2024semi} to train a multimodal global workspace able to translate information and align representations between the vision (image) and digit (one-hot) domains. This is achieved by teaching networks $e_V$ and $e_D$ to encode domain-specific information into the workspace, and networks $d_V$ and $d_D$ to decode workspace information back into the domains (Figure~\ref{fig:architecture_shimmer_2}). We use the four losses introduced by \cite{devillers2024semi}. The contrastive loss serves to align the workspace representations of the image of a digit and that of the digit's identity, that is to align $e_V(\mathcal{E}(\text{Image}(A_R)))$ and $e_D(\text{Label}(A_R))$ (with $(\text{Image}(A_R), \text{Label}(A_R)) \in \text{MNIST}$). We refer the reader to \cite{devillers2024semi} or \cite{radford2021learning} for mathematical details on the contrastive loss. Next is another supervised loss which minimizes the translation error between modalities, that is, the MSE difference between $\text{Label}(A_R)$ and $d_D[e_V(\mathcal{E}(\text{Image}(A_R)))]$, as well as between $\mathcal{E}(\text{Image}(A_R))$ and $d_V[e_D(\text{Label}(A_R))]$. Two unsupervised losses are also used to regularize training by ensuring cycle-consistency. The demi-cycle-consistency loss minimizes the MSE difference between $\text{Label}(A_R)$ and $d_D[e_D(\text{Label}(A_R))]$, and conversely between $\mathcal{E}(\text{Image}(A_R))$ and $d_V[e_V(\mathcal{E}(\text{Image}(A_R)))]$. Finally, the (full) cycle-consistency loss minimizes the MSE discrepancy between $\text{Label}(A_R)$ and $d_D\big[e_V\big(d_V[e_D(\text{Label}(A_R))]\big)\big]$, and between $\mathcal{E}(\text{Image}(A_R))$ and $d_V\big[e_D\big(d_D[e_V(\mathcal{E}(\text{Image}(A_R)))]\big)\big]$; in other words, it ensures consistency after a cycle consisting of a translation followed by the corresponding back-translation. For a justification of these training objectives and a discussion of their respective utility for multimodal representation learning, see \citet{devillers2024semi}.

The global workspace has $10$ dimensions. Its encoders and decoders are composed of 5 feedforward layers with ReLU activation and a hidden size of $32$. They were trained with a learning rate of $1\text{e}^{-3}$ and a batch size of $1024$. The resulting multimodal global workspace is able to translate one-hot labels into meaningful digit images (Supplementary Figure \ref{fig:a2v_translation}). It is also able to cycle through the global workspace and the digit domain without losing meaningful image information (Supplementary Figure \ref{fig:reconstructions}). Furthermore, the digits are better organized (i.e., disentangled) in the global workspace than in the latent space of the autoencoder or in the pixel space, as revealed in Supplementary Figure~\ref{fig:tsne} and confirmed by various cluster metrics (Supplementary Table~\ref{table:cluster_metrics}).

We then freeze and plug the resulting components (encoders and decoders) into the augmented Global Workspace model illustrated in Figure~\ref{fig:architecture}, in lieu of the identity functions. Next, we train the operator module (incrementer) by replacing the router output with the correct ``ground-truth'' gating sequences: first open $g_V$, then open $g_O$ during $A_L \in [1..9]_{\mathbb{N}}$ time steps, then open $g_D$ one time (i.e. here $T=A_L+2$). We construct training $(A_L,A_R) \in [1..9]_{\mathbb{N}}\times \text{MNIST}_\text{training}$ and test $(A_L,A_R) \in [1..9]_{\mathbb{N}}\times \text{MNIST}_\text{test}$ datasets and randomly sample $10,000$ examples from each of them. Here, the incrementer function $o$ is a network of 5 feedforward layers with ReLU activation and a hidden size of $32$; it takes a GW state as input, and returns an updated GW state as output. The incrementer operator was trained with a batch size of $512$ and a learning rate of $5\text{e}^{-4}$ during $1,000$ epochs. Figure~\ref{fig:test_router} (bottom) illustrates that it was able to learn from scratch a function analogous to the hand-designed transition matrix of the one-hot model.

\begin{figure*}[!tb]
\begin{center}
  \includegraphics[clip, trim=0cm 0cm 0cm 0cm, width=1\textwidth]{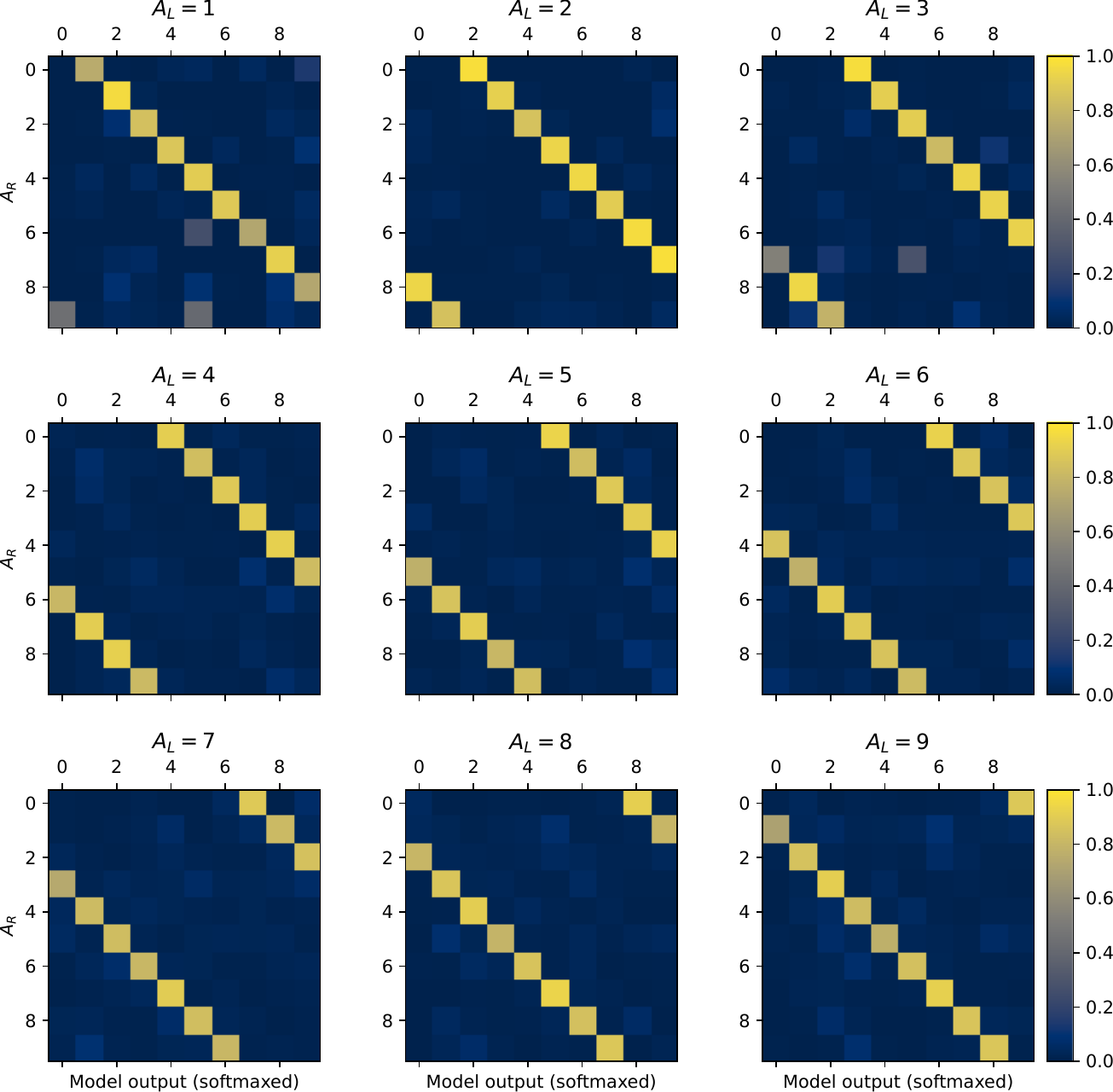}
  \end{center}
    \caption{Behavior of the MNIST model for all combinations of addends $A_L$ (across distinct panels) and $A_R$ (across matrix rows in each panel). The same method as in Figure \ref{fig:test_router} (bottom) was used, except that the model output is shown at the time step where the output gate $g_O$ is maximum, instead of stopping the model at $t=2$. Overall, the model learns to perform the addition tasks correctly.}
\label{fig:test_router_allright}
\end{figure*}

\begin{figure}[!tb]
\begin{center}
  \includegraphics[clip, trim=0cm 0cm 0cm 0cm, width=.48\textwidth]{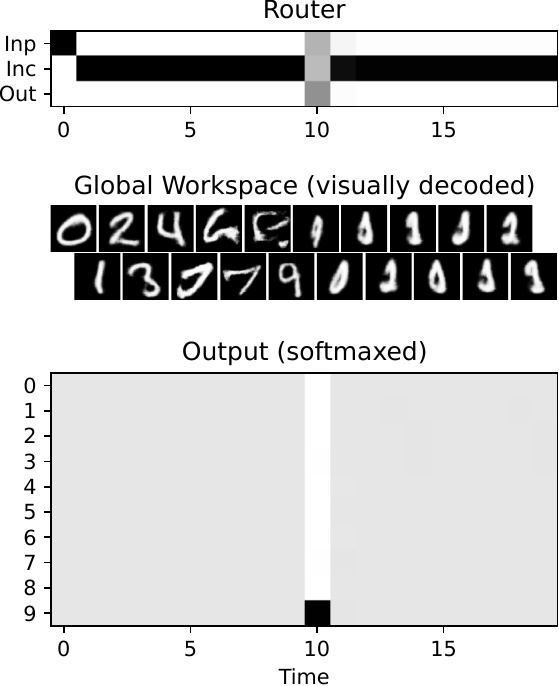}
  \end{center}

    \caption{Behavior of the MNIST model with $A_L=9$ and $A_R=0$. The top panel represents the gating sequence learned by the router. The middle panel represents the global workspace states decoded visually, using $d_V$ and $\mathcal{D}$. The bottom panel shows the activity of the digit domain, softmaxed over the digit dimension. We see the GW state increase from 0 to 9 over the first 10 time steps; at this stage, the answer is routed to the Output module, and the trial is classified as correct. Subsequent time steps reveal ``non-sense'' representations in the GW, which are not detrimental to the task because they are not routed towards the Output.}
\label{fig:mnist_router_states}
\end{figure}

Finally, we freeze $o$ and let the router automatically learn the gating sequences, just as for the one-hot model. We use a LSTM with 2 hidden layers of size 64 that we train with a batch size of $512$, a learning rate of $2\text{e}^{-3}$, a weight decay of $1\text{e}^{-5}$ and a dropout of $2\text{e}^{-1}$ during $10,000$ epochs. Figure~\ref{fig:test_router_allright} shows that, for all combinations of addends, the trained model outputs the correct answer. Inspection of the model's behavior reveals that the router learned to use the incrementer the right number of times (Figure~\ref{fig:mnist_router_states}). The test accuracy remains high ($90.0\%$) albeit slightly lower than for the one-hot model. A noticeable difference between this model's behavior and the one-hot's is that the MNIST router often keeps incrementing after outputting its response. We expect this behavior (which should be penalized by the loss computed over all time steps) to be fixed by optimizing hyperparameters.  

\subsection{Generalization}
In the previous sections, we described a model solving addition tasks by iteratively chaining known operations, in a way akin to System-2 reasoning. This is enabled by its architecture inspired by the Global Workspace Theory. In contrast to System-1 memorization, this compositional way of solving the problem should enable the model to solve unseen tasks. We thus tested generalization to tasks (i.e. values of \( A_L \)) on which the model was not trained. In order to do that, we retrained the model from scratch, omitting some values of $A_L$ from the training set. We divided cases in which the unseen tested value of \( A_L \) was within the bounds of training values (interpolated) and cases in which it was outside the bounds (extrapolated). In order to facilitate comparisons with alternative ``baseline'' architectures, we used the optimized one-hot model and not the MNIST model in this analysis.

For comparison, we consider two alternative models that are conventionally used for sequential tasks: a pure LSTM and a decoder-only causal Transformer. We let the Transformer learn input and positional embeddings, as well as the output projection. We used the transformer implementation of nanoGPT\footnote{https://github.com/karpathy/nanoGPT}. Both models similarly operate during 20 time steps, receiving $A_L$ all along the task and the one-hot representation of $A_R$ at $t=0$. In the Transformer model, the input is positionally embedded. The models output the predicted sum during all 20 steps, without specialized modules nor global workspace. They are trained using the same cross-entropy loss averaged over all 20 time steps as the Global Workspace model.

\begin{figure}[!tb]
\begin{center}
  \includegraphics[clip, trim=0.5cm 0cm 0cm 0cm, width=0.5\textwidth]{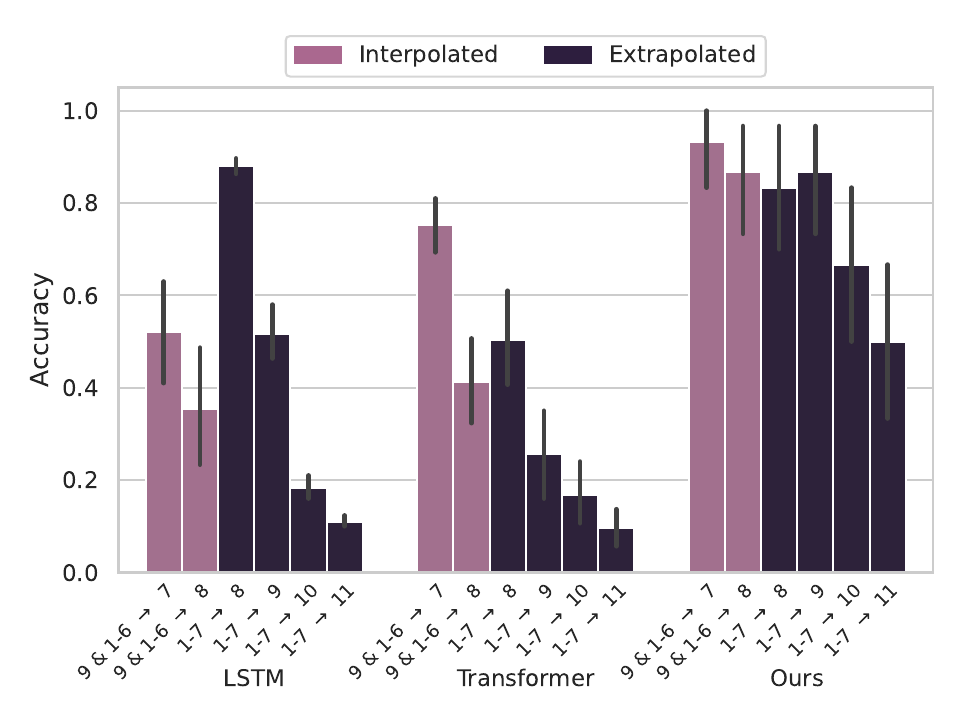}
  \end{center}
    \caption{Performance of the Global Workspace model (right, ``Ours'') and the two baseline architectures (LSTM, Transformer) in the generalization tests. Left-hand side and right-side of ``$\rightarrow$" represent training and test values of \( A_L \) respectively. For instance, $1-7 \rightarrow 9$ indicates that the model was trained to add $A_L$ values from 1 to 7 (with $A_R$ varying from 0 to 9), and subsequently tested on its ability to add $A_L = 9$ (same $A_R$ range). Error bars represent 95\% confidence intervals.}
\label{fig:model_performance_hard}
\end{figure}

After the same hyperparameter optimization as used for the one-hot Global Workspace model (resulting values in Supplementary Table~\ref{table:hyperparameters}), the alternative models are also able to learn the task to some extent (Supplementary Figure \ref{fig:learning_curves}). 
However, the generalization results reveal several key patterns (Figure \ref{fig:model_performance_hard}). While all models can generalize to some unseen tasks, the LSTM and Transformer models struggle with increasing extrapolation. In comparison, our model significantly outperforms others in all but one condition (i.e. when trained on \( [1-7] \) and tested on 8).

\begin{figure}[!tb]
  \begin{center}
  \includegraphics[clip, trim=0cm 0cm 0cm 0cm, width=0.5\textwidth]{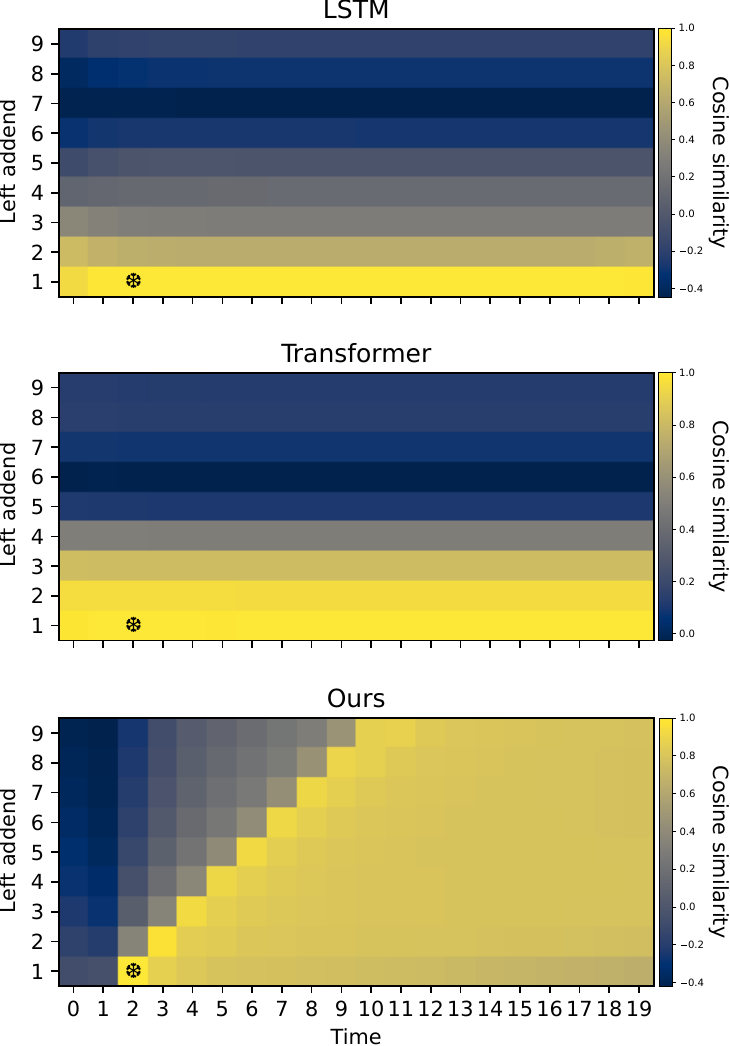}
  \end{center}
    \caption{Cosine similarity between the state of the last hidden layer of the models for each $t$ and $A_L$, and a corresponding reference state ($t=2$ and $A_L=1$) marked with a \ding{100}. $A_R$ is set to $0$. By definition, the similarity at $t=2$ and $A_L$ is $1$ for all models. The states of the LSTM (top) and Transformer (center) models do not evolve over time, but only depend on the number to add. This suggests that these models adopt a System-1 strategy, as they directly reach a memorized state to answer. On the other hand, the state of our Global Workspace model (bottom) depends on both the number to add and time. Most importantly, with increasing $A_L$ values, the model takes a linearly increasing amount of time before reaching a state similar to the reference state. This demonstrates that the router learned to count using a strategy akin to a number line representation.}
\label{fig:similarity}
\end{figure}

In order to understand these differences in generalization abilities across architectures, we investigated the behavior of the candidate models. The internal state of a model solving a task using a multi-step strategy is expected to evolve over time. We predicted that the time required to provide a confident answer during such a strategy should depend on the number to add, $A_L$. We observe this behavior in the Global Workspace model, whose confidence builds over time at a speed depending linearly on \( A_L \) (Supplementary Figure~\ref{fig:confidence_lineplot}). We did not find this in other models. We also predicted that during the iterative integration of $A_L$, a model's internal state at step $t$ should closely resemble its state at step $t+1$ when integrating $A_{L+1}$. This is a hallmark of systematicity, a form of compositionality. We find this pattern clearly in our model (Figure \ref{fig:similarity}), but not in LSTMs or Transformers, whose states do not evolve much over time, and markedly differ for different addition instructions $A_L$.



\section{Discussion}
The Global Workspace Theory provides a compelling framework for explaining human cognition, emphasizing the role of a shared workspace that integrates and broadcasts information from specialized cognitive modules. Inspired by this framework, our model leverages a central workspace and specialized modules to tackle a simple arithmetic task involving sequential, System-2-like reasoning: the model can learn to repeatedly increment the right addend until it reaches the correct sum. The router facilitates this by coordinating the flow of information between the input, incrementer, and output modules, ensuring that the operations are performed in the correct sequence.

We then show that the model can translate information between different representations to coordinate multimodal modules (vision and labels) by leveraging the workspace as a modality-agnostic buffer. The workspace creates disentangled representations of MNIST images that can be used by the modules. We provide a demonstration that the operator module can learn to increment the digit in the workspace representation from scratch, by supervising it distally, on the digit domain side. These encouraging results suggest that the proposed architecture could serve a more general purpose. The model could for instance learn to combine information of multiple sensory modalities (images, sounds, text) and perform a sequence of acquired operations (skills) to solve complex problems, just as humans and other animals do \citep{boraud2018natural}. This can also be related to work on world models, a class of generative neural networks that learn to internally simulate behavioral trajectories and their action on the environment \citep{ha2018world}. In order to reach these more ambitious goals, the relatively simple attentional router used in this work could be extended to open gates according to the content of the global workspace and its specialized modules, as envisioned in \cite{vanrullen2021deep}.

The ability to generalize beyond the training data is a hallmark of intelligent systems and is thought to be enabled by System-2 in humans \citep{kahneman2011thinking, bengio2019system}. We thus wanted to test whether our model could generalize better than more traditional architectures. In our experiments, the Global Workspace model consistently outperformed both LSTMs and Transformers when tasked with generalizing to new instruction (left addend) values that were not part of the training set. This ability is particularly important in extrapolation tasks, where the model must apply learned strategies to out-of-distribution values. While LSTMs and Transformers showed signs of overfitting, failing to generalize to novel values, the Global Workspace model maintained robust performance. We attribute this to the model's explicit representation of sequential reasoning, which enables it to learn a procedure rather than memorizing specific input-output mappings. An investigation of the network dynamics indeed revealed that our model leverages both the temporal and compositional aspects of the task. In contrast, the Transformer and LSTM models primarily learn direct mappings and generate answers from the outset, which limits their generalization capabilities. Furthermore, the sharp learning curves exhibited by our model (Supplementary Figure \ref{fig:learning_curves}) suggest that it quickly grasped the underlying strategy required for solving the task, showing signs of "insight," a sudden leap in understanding the correct approach to chaining operations.

Interestingly, previous work has shown that large pre-trained language models can perform multi-step computation, such as addition, with higher accuracy when accessing an intermediate textual scratchpad \citep{nye2021show}. Our work largely aligns with these results, in that the global workspace can be seen as an internal scratchpad in which the result of intermediate steps can be stored before outputting the final answer. However, recent advances suggest that reasoning in an explicit language space may not always be optimal. For instance, \cite{hao2024training} introduces Coconut, a method in which intermediate reasoning occurs within a continuous latent space, rather than being explicitly represented as text. This approach enhances reasoning efficiency in language models and enables backtracking and parallel exploration of multiple possibilities, similar to how our global workspace operates as an internal, dynamic buffer for multi-step processing.

Finally, we are currently exploring different loss functions that encourage multi-step resolution, which could enrich the comparison with LSTMs and Transformers. As we continue to refine the model and explore more complex tasks, we believe that this approach could serve as a blueprint for future advancements in deep learning and cognitive AI.



\section{Acknowledgments}
This research was funded by an ERC Advanced Grant GLoW (grant 101096017) and an ANITI Chair (ANR grant ANR-19-PI3A-004). 


\bibliographystyle{ccn_style}

\bibliography{ccn_style}

\FloatBarrier

\appendix

\newpage
\onecolumn

\section{Supplementary Material}

\begin{figure}[!h]
\begin{center}
  \includegraphics[clip, trim=0cm 0cm 0cm 0cm, width=.8\textwidth]{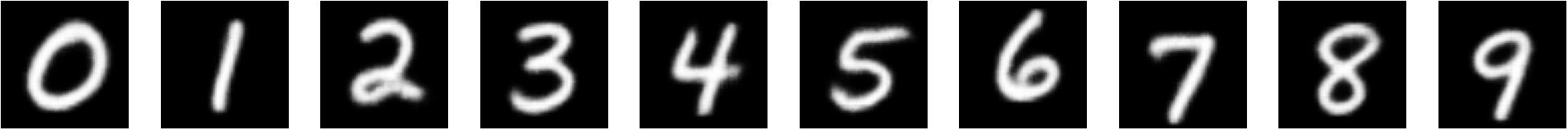}
  \end{center}
    \caption{Decoded translations from the digit domain to the vision domain (i.e. $\mathcal{D}\big(d_V[e_D(\text{Label}(i))]\big), i \in [0-9]$).}
\label{fig:a2v_translation}
\end{figure}

\begin{figure*}[!h]
\begin{center}
  \includegraphics[clip, trim=0cm 0cm 42.5cm 0cm, width=1\textwidth]{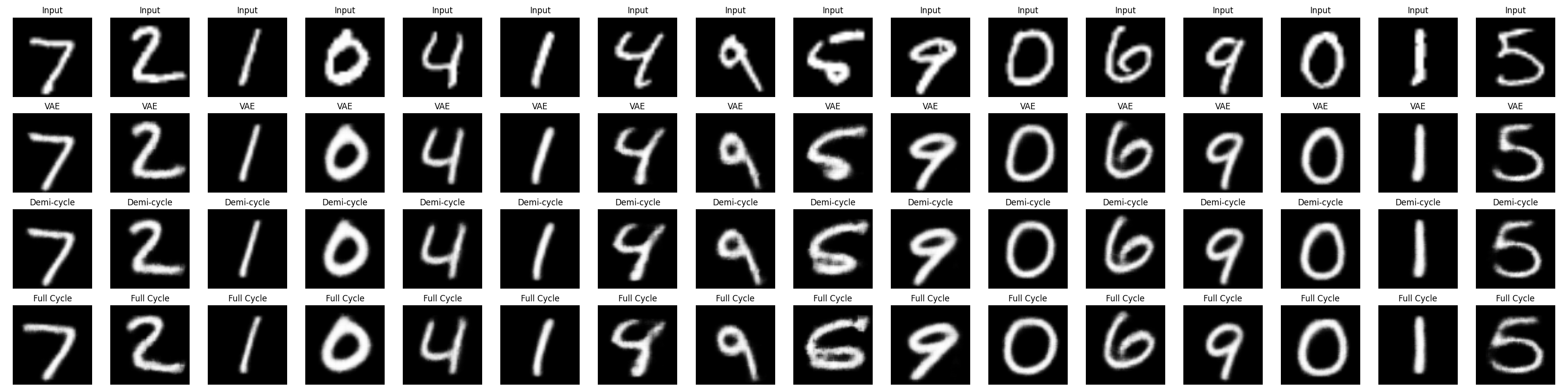}
  \end{center}
    \caption{From top to bottom: raw MNIST images ($\text{Image}(A_R)$); reconstructions by the variational autoencoder ($\mathcal{D}(\mathcal{E}(\text{Image}(A_R)))$); demi-cycles starting from the vision domain, through the Global Workspace and back ($\mathcal{D}(d_V(e_V(\mathcal{E}(\text{Image}(A_R)))))$); full cycles starting from vision, through the GW to the digit domain, and back to the image domain ($\mathcal{D}(d_V(e_D(d_D(e_V(\mathcal{E}(\text{Image}(A_R)))))))$).}
\label{fig:reconstructions}
\end{figure*}

\begin{figure}[!h]
\begin{center}
  \includegraphics[clip, trim=0.4cm .4cm .8cm .35cm, width=0.5\textwidth]{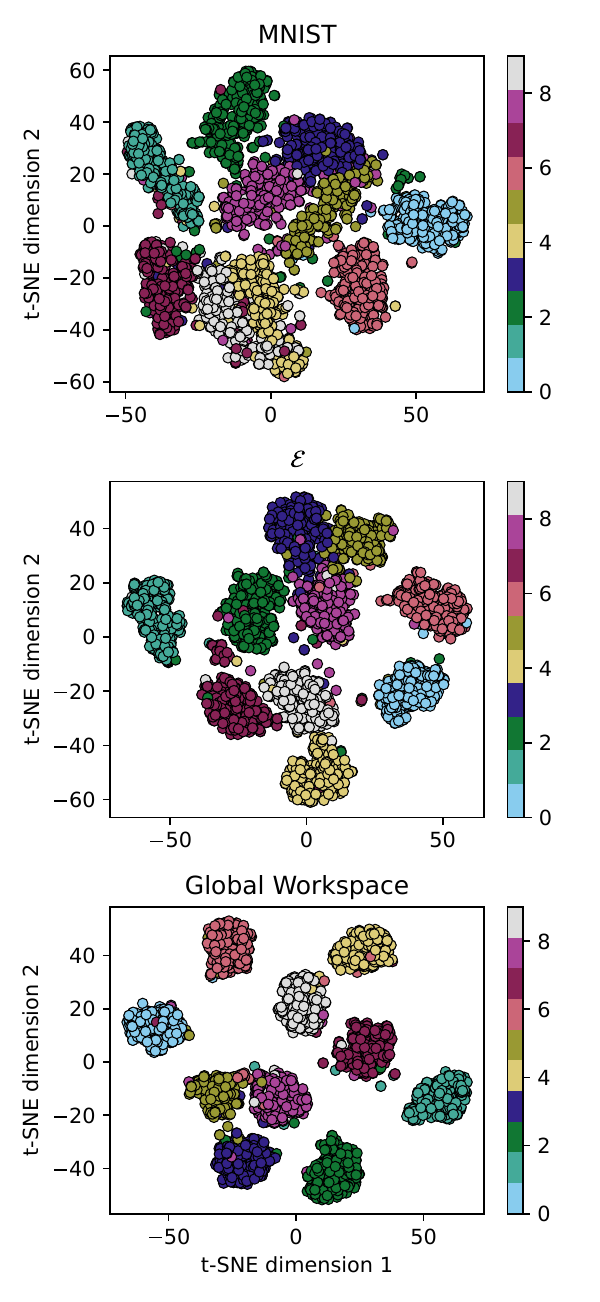}
  \end{center}
    \caption{t-SNE embedding of the MNIST digits, taking as input the pixel values (top), the vision domain states (middle; $\mathcal{E}(\text{Image})$), and the global workspace states (bottom; $e_V(\mathcal{E}(\text{Image}))$).}
\label{fig:tsne}
\end{figure}



\begin{table*}[!h]
\begin{center} 
\vskip 0.12in
\begin{tabular}{|c|c|c|c|}
    \hline
    \textbf{Representation} & \textbf{Silhouette coefficient} & \textbf{Calinski-Harabasz index} & \textbf{Davies-Bouldin index} \\ \hline
    MNIST   & $-0.03$ & $55.48$ & $4.15$  \\ \hline
    $\mathcal{E}$   & $0.15$ & $224.3$ & $1.96$ \\ \hline
    Global Workspace   & $0.23$   & $432.21$ & $1.53$ \\
    \hline
  \end{tabular}
  \end{center} 

  \caption{Clustering metrics of the embeddings illustrated in Supplementary Figure \ref{fig:tsne}. Note that unlike the two other metrics, lower values of the Davies-Bouldin index indicate better clustering.}
  \label{table:cluster_metrics}
\end{table*}

\begin{table*}[!h]
\begin{center} 
\vskip 0.12in
\begin{tabular}{|c|c|c|c|}
    \hline
    \textbf{Parameter} & \textbf{LSTM} & \textbf{Transformer} & \textbf{Global Workspace Router} \\ \hline
    n\_layer   & $4$ & $3$ & $1$  \\ \hline
    hidden\_size   & $96$ & $64$ & $128$ \\ \hline
    lr   & $1\text{e}^{-4}$   & $3.6\text{e}^{-3}$ & $3.5\text{e}^{-3}$   \\ \hline
    weight\_decay   & $2.1\text{e}^{-3}$   & $1.6\text{e}^{-4}$ & $2.5\text{e}^{-4}$   \\ \hline
    dropout   & $0.27$   &  $1.1\text{e}^{-2}$ & $1.6\text{e}^{-2}$ \\ \hline
    n\_head   & -   & $4$ & -   \\ \hline
    bias\_in\_transformer   & - & No & -   \\ 
    \hline
  \end{tabular}
  \end{center} 

  \caption{Hyperparameter values resulting from the bayesian optimization of the three models.}
  \label{table:hyperparameters}
\end{table*}

\begin{figure*}[!h]
  \begin{center}
  \includegraphics[clip, trim=3.7cm 0cm 0cm 0cm, width=1\textwidth]{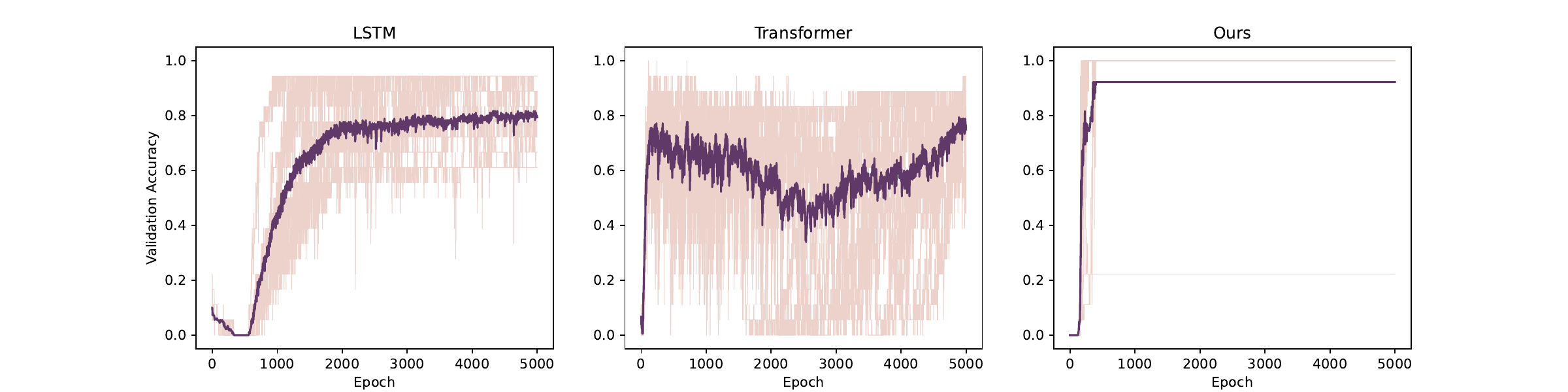}
  \end{center}
    \caption{Learning curves of 10 runs for each candidate model (lighter colors), and their average (darker colors).}
\label{fig:learning_curves}
\end{figure*}

\begin{figure}[!h]
  \begin{center}
  \includegraphics[clip, trim=0cm 0cm 0cm 0cm, width=1\textwidth]{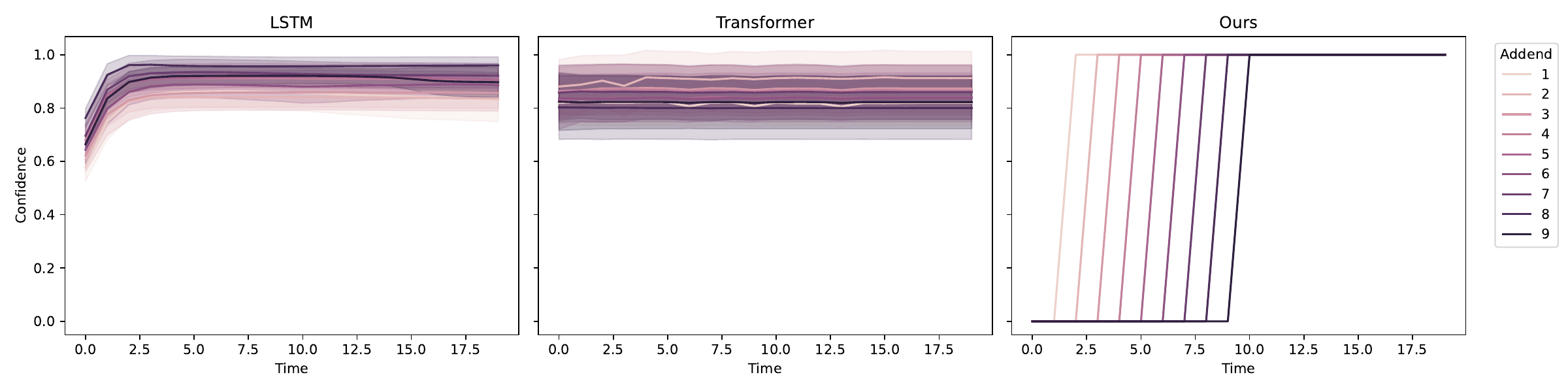}
  \end{center}
    \caption{Normalized confidence (max value of the output vector) as a function of time and \( A_L \) in the successful test samples of 10 runs. For small \( A_L \) values (lighter colors), our model's confidence jumps from 0 to 1 after only a few time steps. As the number to add increases (darker colors), the model takes more time to provide an answer. Contrastingly, the confidence of the other models do not depend much on time and $A_L$.   The shaded region around each line indicates ±1 standard deviation.}
\label{fig:confidence_lineplot}
\end{figure}

\end{document}